\documentclass[sn-mathphys-num]{sn-jnl} 
\usepackage{graphicx}
\usepackage{multirow}
\usepackage{amsmath,amssymb,amsfonts}
\usepackage{amsthm}
\usepackage{mathrsfs}
\usepackage[title]{appendix}
\usepackage{xcolor}
\usepackage{textcomp}
\usepackage{manyfoot}
\usepackage{booktabs}
\usepackage{siunitx}
\usepackage{algorithm}
\usepackage{algorithmicx}
\usepackage{algpseudocode}
\usepackage{listings}
\usepackage{array} 
\usepackage{float} 
\usepackage{lscape}
\usepackage{natbib}
\usepackage{comment}
\usepackage{url}
\usepackage[utf8]{inputenc}

\begin{document}

\title[Article Title]{Towards Sustainable Artificial Intelligence: A Comprehensive Review and Comparative Analysis of Deep Learning Models' Carbon Footprint}
\author*[1]{\fnm{Samar} \sur{Garrab}}\email{samar.garrab@rmc.ca}

\author[2]{\fnm{Sarra} \sur{Boughriou}}\email{sarra.boughriou@mail.concordia.ca}

\author[3]{\fnm{Manel} \sur{Ben Sassi}}\email{manel.bensassi@ensi-uma.tn}

\affil*[1]{\orgname{Royal Military College of Canada},\city{ Ontario},\country{ Canada}}
\affil[2]{\orgname{Concordia University},\city{ Quebec},\country{ Canada}}

\affil[3]{\orgname{École Nationale des Sciences de l’Informatique},\country{ Tunisia}}
\abstract{
Artificial Intelligence (AI) and Machine Learning (ML) have become powerful tools for supporting and automating complex human tasks. Despite their benefits, growing attention has been directed toward their environmental implications, primarily due to their high energy demands and associated carbon emissions. This concern is particularly relevant in light of the increasing deployment of large-scale models, especially Deep Learning (DL) architectures, which provide advanced predictive capabilities but require substantial computational resources.

This paper presents a systematic review of research on Green AI, Green DL, and optimization techniques aimed at reducing the environmental impact of AI models. In addition, we examine and compare several carbon measurement tools for estimating emissions generated by AI algorithms. 

To complement the review, we conducted an empirical evaluation using a CPU-based experimental setup, in which six DL models were implemented for a multi-label classification task. The objective was to quantify and compare their overall carbon emissions and to determine which stages of the DL lifecycle contribute most significantly to the total footprint. The results show that the training phase is the primary source of emissions. Moreover, the findings reveal that increased architectural complexity does not systematically translate into proportional accuracy gains, highlighting the importance of carefully balancing predictive performance and environmental cost. These results reinforce the need to integrate sustainability considerations into model selection and AI system design.
}
\keywords{Green Artificial Intelligence, Green Deep Learning, Carbon footprint assessment tools, Sustainable development, Model optimization, Resilient algorithms.}
\maketitle
\section{Introduction}\label{sec1}

Sustainable development refers to fulfilling present needs without undermining the ability of future generations to meet their own. It is traditionally structured around three interconnected pillars: economic prosperity, social equity, and environmental stewardship. In this context, Artificial Intelligence (AI) and Machine Learning (ML) have emerged as transformative technologies with the potential to accelerate progress toward the Sustainable Development Goals. Their applications span a wide range of global challenges, including energy optimization, climate modeling, and efficient resource management~\cite{verdecchia2023systematic,rolnick2019tackling}.

Nevertheless, the interaction between AI and sustainability is inherently paradoxical. While AI systems can contribute to environmental solutions, the processes involved in developing, training, and deploying these systems generate substantial environmental costs. In particular, large-scale models require considerable computational power, resulting in high energy consumption and associated carbon emissions~\cite{strubell2019energy,henderson2020towards}. This duality raises a critical concern: how can society harness the benefits of AI while minimizing its ecological footprint?

The complexity of this issue lies in the fact that AI simultaneously serves as both a mitigation tool and a source of environmental pressure. As discussed in~\cite{pachot2022towards}, AI technologies may support climate action initiatives while also intensifying the very environmental challenges they aim to address. Similarly, Dhar~\cite{dhar2020carbon} characterizes AI as both a promising ally in the pursuit of sustainability and a concealed contributor to climate change due to its growing carbon footprint.

Deep Learning (DL), a prominent subfield of AI, exemplifies this tension. DL techniques are increasingly being adopted across diverse domains, including healthcare, finance, autonomous systems, and natural language processing. Continuous methodological advancements and the pursuit of state-of-the-art performance have led researchers to rely on progressively larger models and more powerful hardware infrastructures. However, these performance gains often come at the expense of substantial energy requirements, particularly during the training phase of DL models.

A fundamental step toward mitigating the environmental impact of AI is the accurate measurement of its energy use and carbon emissions. Quantifying these factors is essential for understanding and ultimately reducing the environmental burden of AI systems. Prior studies have emphasized that AI models can be energy-intensive throughout their entire lifecycle, including design, training, deployment, and operational phases~\cite{schwartz2020green}.

Motivated by these concerns, this paper seeks to provide a clearer and more rigorous assessment of the environmental implications of AI algorithms, with a particular emphasis on Deep Learning. While existing literature reviews address Green AI from a broad perspective~\cite{verdecchia2023systematic} and discuss Green DL primarily at the conceptual or methodological level~\cite{xu2021survey}, they often lack reproducible quantitative evaluations that compare the carbon footprint of DL models across distinct lifecycle stages. This gap underscores the need for systematic, experimentally grounded analyses, which our work aims to provide.

In this paper, we conducted a literature review covering Green Artificial Intelligence (Green AI), Green Deep Learning (Green DL), and existing tools for measuring carbon footprints. We then performed a controlled empirical study comparing the carbon footprints of six widely used deep learning architectures— simple Convolutional Neural Network (CNN), Residual Networks (ResNet), U-shaped Network (U-Net), Visual Geometry Group networks with 16 and 19 layers (VGG16 and VGG19), and Efficient Networks (EfficientNet)—and analyzed their emissions both overall and across individual phases, namely data preprocessing, model training, and testing. To further quantify the relationship between predictive performance and environmental impact, we compute the Emissions per Accuracy Point (EAP), a carbon-efficiency metric that relates total emissions to achieved accuracy, thereby enabling direct comparison of the sustainability of different models.

This paper goes beyond prior studies in three main aspects. First, we conduct a Systematic Literature Review (SLR) to provide a structured comparative analysis of existing carbon footprint measurement tools, discussing their respective strengths and limitations. Second, we introduce a fully reproducible experimental setup using CodeCarbon to measure carbon emissions across the different phases of the deep learning lifecycle. Third, we present a benchmark of six representative deep learning architectures that jointly report predictive performance and per-phase carbon emissions.

Section~\ref{sec:Littrev} provides a foundational review of relevant research on Green AI, identified through an SLR and mapping methodology. Building on this foundation, Section~\ref{sec:carb} examines and compares existing tools for measuring the carbon footprint of AI models. To gain practical insights, Section~\ref{sec:Compt} presents a computational analysis of deep learning models, describing the selected models, the experimental environment, and the data exploration procedures.

Section~\ref{sec:results} quantifies the carbon footprint of the evaluated DL models, both overall and by individual lifecycle phases, in order to pinpoint the stages that contribute most to emissions. Following this, Section~\ref{sec:discussion} explores the broader implications of these results, emphasizing the trade-offs between model performance and environmental sustainability. Section~\ref{sec:threatsValidity} addresses potential threats to the validity of the study.

Finally, Section~\ref{sec:conc} concludes the article by summarizing the key findings and outlining directions for future research and development in sustainable AI practices.

\section{Literature review}
\label{sec:Littrev}
The existing literature on Green AI reveals that several proposals have been advanced, including optimization of DL and AI algorithms. The following section provides an overview of key studies on the development of Green AI models and the implementation of algorithmic optimization techniques. It begins by outlining the research methodology, then defines Green AI and Green DL and discusses their associated optimization approaches.

\subsection{Research Methodology}
This review investigates the current state of research on Green AI and Green DL, with particular emphasis on studies that quantify the carbon footprint of AI algorithms. To achieve this, we conducted a systematic literature review (SLR) and mapping study following the Preferred Reporting Items for Systematic Reviews and Meta-Analyses (PRISMA) framework~\cite{Prisma2021}. This approach provides a structured and comprehensive analysis of the literature, enabling the identification of key trends, research gaps, and the overall landscape of Green AI research.

The mapping study addresses the following mapping questions (MQs):  

\begin{itemize}
    \item[] \textbf{MQ1:} How many studies on Green AI or Green DL have been published over the past ten years (2015–2025)?
    \item[] \textbf{MQ2:} Which tools are commonly employed to measure the carbon footprint of AI algorithms within the context of Green AI?
\end{itemize}

The research questions (RQs) guiding this review are as follows:  

\begin{itemize}
    \item[] \textbf{RQ1:} What strategies have been proposed to enhance the sustainability of AI technologies by reducing their carbon footprint, and how effective are these strategies in mitigating the environmental impact of AI algorithms?
    \item[] \textbf{RQ2:} What tools currently exist for quantifying the carbon footprint of AI algorithms?
\end{itemize}

By systematically addressing these questions, this review aims to synthesize the existing knowledge on Green AI, focusing on studies that assess the environmental impact of AI systems and providing detailed insights into their methodologies, findings, and implications for sustainable development.

To gather relevant literature, we used the following databases: IEEE Xplore, ACM Digital Library, and Scopus. Semantic Scholar was also included as a complementary AI-focused search engine.
We formulate and use the following search queries to identify studies for the past 10 years (2015-2025): 
\begin{itemize}
    \item “Green AI” OR “Green Artificial Intelligence”
    \item “Green DL” OR “Green Deep Learning”
    \item “Carbon footprint tools”
    
\end{itemize}

All retrieved records were manually screened to remove duplicates and non-relevant studies, ensuring both relevance and quality of the final selection.
 Table~\ref{tab:slr} illustrates the number of studies found per topic in each database.
\begin{table}[htbp]
\caption{Number of articles retrieved per topic and per database}\label{tab:slr}
\begin{tabular}{@{}p{3cm}p{2.5cm}p{2.5cm}p{2.4cm}p{0.8cm}@{}}
\toprule
Database & Green Artificial Intelligence/ Green AI  & Green Deep Learning/ Green DL & Carbon Footprint Tools & Total \\\midrule
IEEE Xplore  & 137 & 2 & 2 &  141 \\
ACM Digital Library & 84 & 3 & 4 & 91 \\ 
Scopus & 283 & 7 & 7  & 297 \\
Semantic Scholar  & 519  & 40 &  72 & 631 \\
\bottomrule
\end{tabular}
\end{table}

\begin{center}
\begin{figure}[!htpb]
\centering
\includegraphics[width=0.9\textwidth]{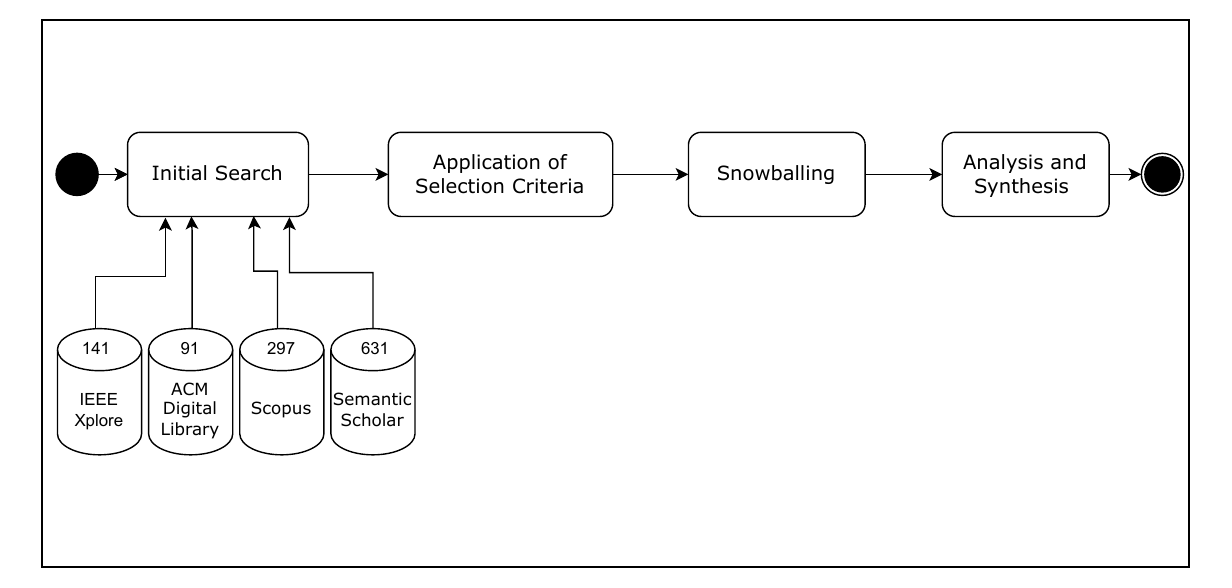}
    \caption{Overview of the systematic literature review (SLR) process adopted in this study}
    \label{fig:slr}
\end{figure}

\end{center}

Following the identification of potentially relevant studies, we carried out the systematic literature review (SLR) process, as depicted in Fig.~\ref{fig:slr}.

We manually selected studies according to predefined selection criteria~\cite{verdecchia2023systematic}.  
The following inclusion (IC) and exclusion (EC) criteria were applied: \\

\textbf{IC1:} The study is about Green AI or Green DL.  

\textbf{IC2:} The study relates to the measurement of carbon footprint.

\textbf{IC3:} The study must be a peer-reviewed journal article or conference paper.\\

\textbf{EC1:} The study is a duplicate or extension of an already included study.

\textbf{EC2:} The study is not written in English.

\textbf{EC3:} The study is not available.

\textbf{EC4:} The study is not related to computer science.

\textbf{EC5:} The study is not relevant to the research questions.

Selected studies were original peer-reviewed journal articles or conference papers (IC3) that focus on Green AI or Green DL (IC1) or address carbon footprint measurement (IC2). They also did not meet any of the exclusion criteria (EC1–EC5).

\begin{figure}[h]
    \centering
    \includegraphics[width=0.9\linewidth]{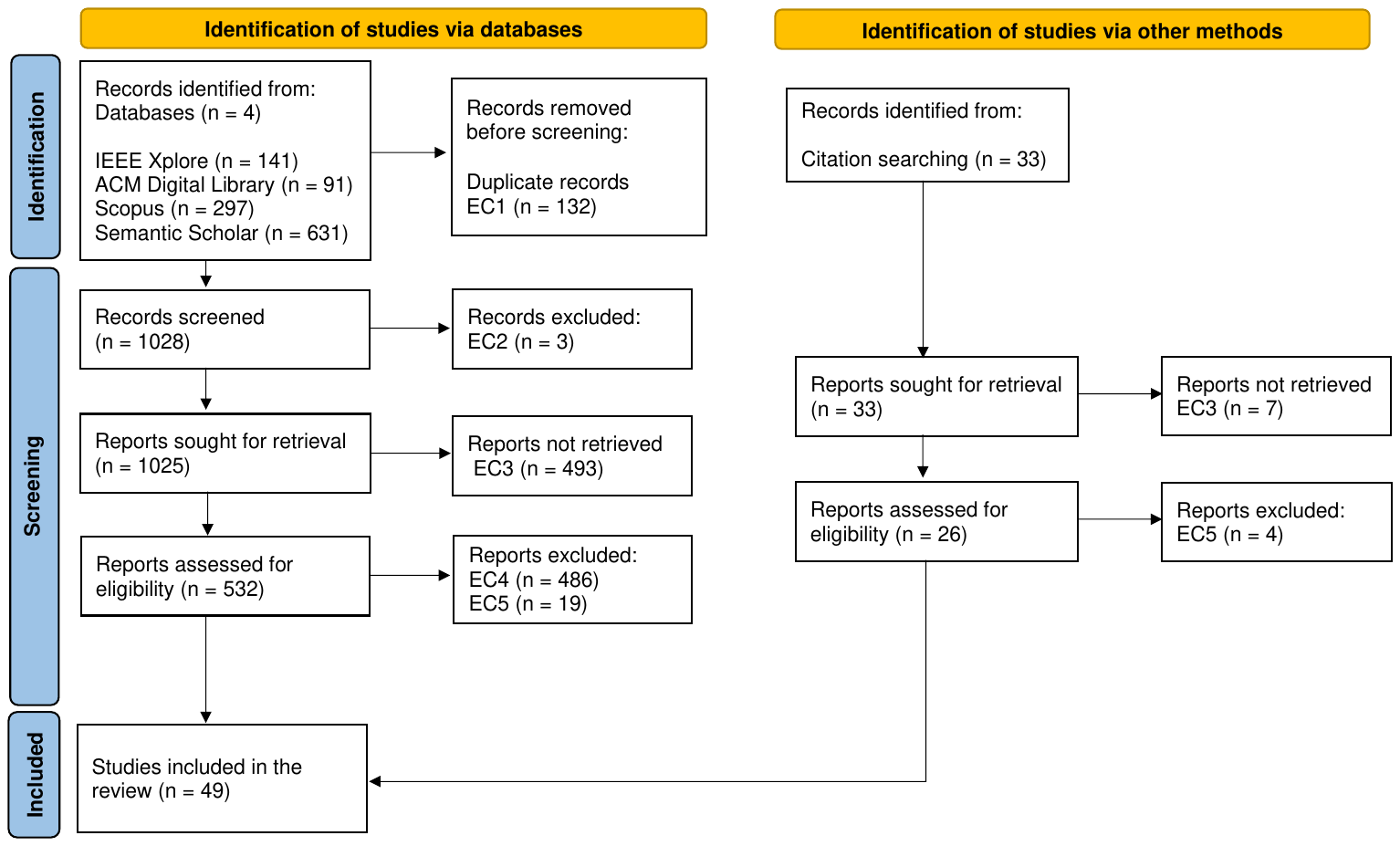} \vspace{-1.5cm}
    \caption{PRISMA flow diagram of the study selection process}
    \label{fig:screening}
\end{figure}

Fig.~\ref{fig:screening} provides a detailed PRISMA flow diagram of the systematic review process.

To ensure comprehensive coverage, we complement the database search with snowballing~\cite{snowballing}, an iterative method that expands the initial set of studies (“seed”) by following citation links. In particular, we applied both \textit{backward snowballing} (examining the reference lists of included papers) and \textit{forward snowballing} (identifying subsequent papers that cite them). This dual approach helps uncover studies that may have been missed by keyword-based searches. In total, these steps resulted in the inclusion of 22 additional articles.

Articles that successfully passed the initial filtering and snowballing stages underwent a thorough full-text review. During this phase, each article was carefully examined to verify its relevance and alignment with the research objectives. This detailed assessment ensured that only the most pertinent and high-quality studies were retained for the final selection.

After finalizing the set of included articles, we conducted a comprehensive analysis and synthesis of the extracted data. Key information was collected from each study, including the reported environmental impact of AI, the methodologies employed for carbon footprint estimation, and the optimization strategies implemented.

Based on this analysis, we addressed the predefined research questions. This systematic process provided a clear picture of the current state of Green AI and Green DL research, highlighting both the challenges and the strategies for measuring and mitigating the carbon footprint of AI algorithms, as discussed in the subsequent subsections.

\subsection{Green AI: Definition and Approaches}
In this section, we provide a definition of Green AI and review its key optimization strategies.

\subsubsection{Definition}
Recent works highlight a rapidly growing body of research on Green AI, developed in response to the increasing environmental footprint of artificial intelligence. Verdecchia et al.~\cite{verdecchia2023systematic} note that studies aimed at improving the energy efficiency of AI models are increasing, reflecting growing concern that AI’s environmental cost may ultimately affect human well-being. 

The notion of “sustainability of AI,” introduced by Wynsberghe~\cite{wynsberghe2021sustainable}, frames it as a concept that seeks change across the entire lifecycle of AI products. Natarajan et al.~\cite{natarajan2022theoretical} expand this perspective by highlighting the need to minimize the environmental impact at every stage—design, training, and inference—through techniques such as model pruning, efficient learning algorithms, and specialized hardware. 

A central distinction in this debate is between “Red AI” and “Green AI.” Red AI, as described by Schwartz et al.~\cite{schwartz2020green}, requires lengthy training, massive datasets, and energy-intensive methods, leading to high emissions and reliance on non-renewable energy. By contrast, Green AI models aim to achieve comparable results with lower energy consumption, smaller datasets, and more efficient training strategies, often leveraging renewable energy sources to further reduce carbon costs~\cite{barbierato2024towards}. This shift is already visible in industry practices: Google has committed to operating entirely on 24/7 carbon-free energy by 2030~\cite{google2024sustainability}, including partnerships to expand geothermal power in Nevada~\cite{groom2024google}, while Microsoft has secured hundreds of megawatts of solar capacity to meet its growing data center demand~\cite{dcd2024microsoft}. These developments underscore that Green AI extends beyond algorithmic optimization and must also be embedded within sustainable energy infrastructures. 

The vision articulated by Schwartz et al.~\cite{schwartz2020green} provides a roadmap for addressing the challenges posed by Red AI, focusing on maximizing performance under resource constraints while minimizing computational costs with minimal sacrifice in accuracy. In this sense, Green AI represents not only a technical agenda but also a pathway to AI practices that benefit society more holistically. Translating this vision into practice requires continued collaboration between academia and industry to ensure that advances in efficiency contribute to both technological progress and environmental responsibility. 

Building on this efficiency-oriented perspective, an emerging paradigm aligned with Green AI is zero-waste AI, which emphasizes eliminating unnecessary computation and reusing computational resources to reduce the environmental impact of AI models~\cite{trzcinski2024zero}. In this sense, zero-waste AI can be viewed as an operational extension of Green AI principles, focusing on preventing avoidable computational waste rather than solely improving energy efficiency.

Finally, recent contributions broaden the scope of Green AI to encompass societal and policy dimensions. Acosta-Vargas et al.~\cite{acosta2024generative} highlight how generative AI can support an inclusive and sustainable future by improving accessibility. Sookhom et al.~\cite{sookhom2023tourism} situate digital innovation within the green economy, showing how it can empower communities and foster sustainable development. Shaari et al.~\cite{shaari2024carbon} examine the links between CO$_2$ emissions, public debt, and environmental policy, underscoring the need to align technological innovation with macroeconomic sustainability strategies. Taken together, these perspectives demonstrate that Green AI is not only about efficiency but also about inclusivity, equity, and responsible governance.

\subsubsection{Multi-metric Sustainability and Trade-offs}
While carbon emissions are a central and widely used indicator of the environmental impact of AI systems, they do not capture the full complexity of sustainability. Several studies have noted that AI sustainability is inherently multidimensional and involves additional factors, including energy consumption, hardware lifetime, storage requirements, and electronic waste generation~\cite{henderson2020towards, wu2022sustainable, verdecchia2023systematic}.

In practice, improving one aspect of sustainability may degrade another. For example, aggressive model compression or repeated architecture search can reduce the emissions of a single training run, but may increase the number of experiments or accelerate hardware aging, thereby shifting part of the environmental burden to the hardware lifecycle~\cite{strubell2019energy, wu2022sustainable}. Conversely, larger or more general models may increase inference-time emissions while reducing the need to maintain and retrain multiple specialized models~\cite{henderson2020towards}.

In addition, infrastructure-level factors such as data center cooling and the production and disposal of specialized accelerators, "Graphics Processing Units (GPUs) and Tensor Processing Units (TPUs)", introduce water consumption and electronic waste concerns, which are typically studied using life-cycle assessment methodologies rather than software-level measurement tools~\cite{lacoste2019quantifying, wu2022sustainable}. These aspects are therefore outside the scope of the present experimental setup.

From a governance perspective, several authors argue that carbon emissions and energy use should be reported alongside traditional performance metrics to enable more responsible and transparent model comparisons~\cite {strubell2019energy, henderson2020towards, wu2022sustainable}. In this work, we focus on operational carbon emissions as a practical and reproducible proxy for environmental impact; however, our results should be interpreted as only one component of a broader sustainability assessment.

\subsubsection{Green AI Optimization Techniques}

Prior research distinguishes between two main categories of optimization strategies for Green AI: hardware-based and software-based approaches~\cite{zhou2023opportunities}.  

\begin{itemize}
    \item \textbf{Hardware optimization} focuses on improving the physical components of the system. For instance, Lannelongue et al.~\cite{lannelongue2021green} highlight the importance of optimizing algorithms to enhance computational speed and reduce memory usage. Similarly, Lacoste et al.~\cite{lacoste2019quantifying} advocate for designing energy-efficient model architectures and employing hardware with lower energy consumption, such as GPUs and TPUs. Both studies underscore strategies to minimize training duration through hyperparameter tuning, advanced algorithmic improvements, and transfer learning, while also emphasizing the use of renewable energy and enhanced data center efficiency.
    \item \textbf{Software optimization} involves refining the code and algorithms of AI models. These adjustments can be applied across different stages of the model lifecycle, including preprocessing, training, and inference, to reduce computational overhead and energy usage.
\end{itemize}

In this work, we focus primarily on software optimization to improve the environmental efficiency of deep learning models. This approach seeks to make models run more efficiently on existing hardware, offering a more accessible and widely applicable solution compared to hardware upgrades.

In Tables~\ref {tab:prep}, \ref {tab:train}, and \ref {tab:inf}, we review the optimization approaches available for each phase.     
\begin{table}[h]
\caption{Optimization techniques used in the preprocessing phase} 
\label{tab:prep}%
\begin{tabular}{@{}p{2.6cm}p{4.8cm}p{4.8cm}@{}}
\toprule

Techniques & Advantages & Disadvantages \\
\midrule

Data Preparation & 
- Improve data quality. 
&
- Consume significant time (45–90\%
 \\
 \cite{whang2023data}
 & 
- Reduce errors in downstream tasks.
 & of project time).
\\
 & 
- Ensure robust models. 
& - Require domain expertise.
 \\
\midrule

Normalization/ Standardization & 
- Ensure features are on the same scale. &
- Distort interpretability. 
 \\
 \cite{ali2014data, cabello2023impact}
 & 
- Improve training stability.
 &
- Require careful choice of method. \\
\midrule
Feature Selection & 
- Reduce dataset size. &
- Discard useful features. \\
\cite{mwangi2014review}
& 
- Lower accuracy. \\

\midrule

Dimensionality & 
- Capture key variance in fewer &
- Reduce interpretability of features.\\
 Projection~\cite{van2009dimensionality} & 
 dimensions. &
 - Remain sensitive to data distribution \\
 (e.g., PCA)
 &
- Reduce computational cost. &
tion.
\\

\midrule

Data Augmentation \cite{hernandez2018further, wang2025comprehensive} & 
- Expand dataset artificially (oversampling, transformations).
 &
- Introduce unrealistic or biased samples. 
\\

& 
- Improve generalization and reduce overfitting. &
- Increase preprocessing time. \\
\bottomrule
\end{tabular}
\end{table}

\begin{table}[tbph]
\caption{Optimization techniques used in the training phase}
\label{tab:train}%
\begin{tabular}{@{}p{2.5cm}p{4.5cm}p{5.1cm}@{}}
\toprule
Techniques & Advantages & Disadvantages \\
\midrule
Initialization& 
- Normalize variance of activations.
 &
- Hinder learning if the scheme is wrong. 
 \\
 \cite{xu2021survey,zhang2019fixup,mishkin2015all}& 
- Improve convergence using pre-trained weights. &
  
- Fail to transfer when pre-trained weights are mismatched. \\
\midrule
Normalization & 
- Stabilize and accelerate training. 
 &
- Add computational overhead.  
\\
 \cite{ioffe2015batch,ba2016layer,wu2018group} & 
- Reduce sensitivity to initialization. &
- Depend on batch size for effectiveness.\\
\midrule
Progressive \hspace{1cm} Training~\cite{belilovsky2019greedy} & 
- Enhance efficiency through curriculum learning.
 &
- Require careful scheduling.
\\
 & 
- Improve performance via progressive resizing. &

- Increase pipeline complexity. \\

\midrule
Hyper-parameter & 
- Improve model accuracy. 
 &
- Consume resources due to extensive   
 \\
Optimization~\cite{yu2018fast}& 
- Save resources through early & search. \\
&stopping. & - Fail to generalize best settings. \\
\bottomrule
\end{tabular}
\end{table}

\begin{table}[h]
\caption{Optimization techniques used in the inference phase}
\label{tab:inf}%
\begin{tabular}{@{}p{2.5cm}p{4.8cm}p{4.8cm}@{}}
\toprule
Techniques & Advantages & Disadvantages \\
\midrule
Pruning~\cite{lecun1990optimal,menghani2023efficient} & 
- Remove redundant weights to 
 &
- Require fine-tuning. 
 \\
 &
 shrink model size.
 &
 - Degrade accuracy if pruning is too
 \\
 & 
- Maintain accuracy with fine-tuning. 
& 
aggressive.
 \\

\midrule
Low-rank \hspace{1cm} Factorization \cite{hsu2022language} & 
- Decompose weights (SVD, CP) to reduce computation. &
- Increase computation cost during decomposition.
 \\
 & 
- Reduce storage costs. &
- Slightly reduce accuracy.\\
\midrule
Quantization~\cite{raghuraman2018quantizing} & 
- Reduce memory footprint by lowering bit precision.
 &
- Depend on bit-width choice.
\\
 & 
- Provide speedups on specialized hardware. &
- Harm accuracy if tuning is poor. \\

\midrule
Knowledge \hspace{1cm}Distillation~\cite{hinton2015distilling,ba2014deep}& 
- Train compact student models with minimal accuracy loss.
 &
- Depend on the teacher–student setup.
\\
& 
- Provide flexibility (logits-, feature-, or relation- based). &
- Require an additional training step.\\
\bottomrule
\end{tabular}
\end{table}

\subsection{Green DL: Definition and Approaches}
In this section, we define Green DL and review several associated optimization strategies.

\subsubsection{Definition}
Deep Learning, a prominent subfield of AI, is distinguished by its complexity and reliance on large-scale artificial neural networks. Unlike traditional AI methods, DL models require extensive datasets and substantial computational resources~\cite{LecunDL}. This growing complexity carries significant environmental implications, linking Green DL closely to the broader concept of Green AI.

Schwartz et al.~\cite{schwartz2020green} highlight the rapid increase in data volume, model parameters, and system demands associated with DL, which contributes to a considerable carbon footprint. Notably, they observe that the inference phase—when the model generates predictions—consumes a larger share of energy than the training phase. Supporting this, Wu et al.~\cite{wu2022sustainable} report that inference accounts for approximately 65\% of a model’s total carbon footprint, compared with 35\% for training. These findings suggest that quantifying carbon emissions alone is insufficient; cost-benefit analyses considering model performance are also essential~\cite{strubell2019energy}.

Further investigations by Li et al.~\cite{li2016evaluating} examine DL energy consumption for tasks such as image classification, emphasizing the need to optimize the entire machine learning pipeline. This perspective is echoed by Henderson et al.~\cite{henderson2020towards}, who stress that environmental efficiency must be considered at every stage of model development and deployment.

\subsubsection{DL Algorithms Optimization }

Optimizing deep learning models is a central aspect of Green AI. Gutiérrez et al.~\cite{gutierrez2022analysing} highlight that even modest adjustments in model design, such as selecting more efficient algorithms, can substantially reduce a model’s environmental footprint.  

However, achieving truly sustainable DL requires a holistic approach. Henderson et al.~\cite{henderson2020towards} emphasize that the entire AI pipeline—from data collection and exploration to training, optimization, and deployment—must be considered. They propose a framework for monitoring energy consumption and carbon emissions throughout the AI lifecycle, demonstrating how comprehensive Green AI practices can ensure that AI development promotes both technological progress and environmental sustainability.  

In a similar vein, Rafiullah~\cite{rafiullah2023ai} stresses the importance of incorporating energy constraints into model design. Their work illustrates the trade-offs between accuracy and efficiency, recommending strategies such as minimizing computational operations, reducing dataset sizes, and eliminating non-informative or redundant features. Such measures can significantly lower energy use during inference without severely compromising model performance.  

By applying these optimization strategies, DL models can achieve strong predictive capabilities while minimizing their environmental impact. To quantify this impact, computational tools are essential, which we discuss in detail in the following section.

 \section{Carbon Footprint Tools}
\label{sec:carb}

This section introduces existing tools for assessing the carbon footprint of AI models and provides a comparative analysis using multiple evaluation criteria to identify the most suitable tool for our study.

\subsection{Measurement Tools}

The open-source community has developed numerous tools for estimating the environmental impact of AI systems. Several of these tools have been highlighted in the literature~\cite{bannour2021evaluating,budennyy2022eco2AI}. We identified seven notable tools, which can be grouped into four categories:

\begin{itemize}
    \item \textbf{Online Calculators:} Green Algorithms~\cite{lannelongue2021green} and ML CO2 Impact~\cite{lacoste2019quantifying} are web-based calculators that estimate the environmental impact of a user’s computational choices. They account for factors such as runtime, number of cores, memory allocation, hardware type, and geographic location.
    \item \textbf{Experiment Measurement Tools:} Tools like Energy Usage~\cite{lottick2019energy} and Experiment Impact Tracker~\cite{henderson2020towards} are designed to help researchers quantify the environmental footprint of ML experiments. They track energy consumption by monitoring hardware utilization, including CPU, GPU, and DRAM usage, typically within a Python environment.
    \item \textbf{Advanced Prediction Tools:} Carbon Tracker~\cite{anthony2020carbontracker} offers predictive capabilities for estimating the energy consumption of DL models during training. It integrates runtime metrics, GPU workload, and carbon intensity data to provide more sophisticated forecasts.
    \item \textbf{Python Packages:} Several Python libraries enable seamless integration of carbon footprint estimation into ML workflows. Examples include CodeCarbon~\cite{codecarbon}, which measures energy consumption and carbon emissions in real time during code execution, and Cumulator~\cite{trebaol2020cumulator}, which estimates energy use based on runtime, hardware specifications, and predefined values.
\end{itemize}

We conducted a comparative analysis of these tools to highlight their distinctions and practical suitability, as discussed in the following subsection.

\subsection{Comparative Study}

To evaluate carbon footprint measurement tools for AI projects, we established a set of key criteria emphasizing reusability, usability, and accuracy.  

Our comparative analysis employed six criteria:  

\noindent \textbf{C1 – Open-source:} Determines whether the tool’s source code is publicly accessible for inspection, modification, and community contribution.  

\noindent \textbf{C2 – Documentation:} Assesses the availability of detailed instructions, including technical guides for installation and usage.  

\noindent \textbf{C3 – Ease of use:} Evaluates the technical expertise required to operate the tool effectively.  

\noindent \textbf{C4 – Method of use:} Describes how the tool is accessed or integrated (e.g., web platform, code injection, desktop application).  

\noindent \textbf{C5 – Functionality:} Categorizes the tool’s output as an estimate, measurement, or prediction of carbon emissions.  

\noindent \textbf{C6 – Measurement detail:} Distinguishes between global assessments, which calculate overall energy consumption or emissions for the entire ML pipeline, and per-phase assessments, which provide estimates for individual stages such as data preprocessing, model training, evaluation, hyperparameter tuning, and inference.  

Using these criteria, we compared seven tools: Green Algorithms, ML CO2 Impact, Energy Usage, Experiment Impact Tracker, Carbon Tracker, CodeCarbon, and Cumulator. Key observations include:  

\begin{itemize}
    \item \textbf{Open-source:} All seven tools are open-source, promoting transparency, collaboration, and continuous improvement. Publicly available code allows users to inspect, modify, and contribute enhancements, fostering reliability and innovation.
    
    \item \textbf{Documentation:} Effective documentation is critical for adoption. While Green Algorithms, ML CO2 Impact, Energy Usage, and Experiment Impact Tracker provide limited guidance, Carbon Tracker, CodeCarbon, and Cumulator offer comprehensive documentation, including clear installation instructions, usage examples, and technical guides.
    
    \item \textbf{Ease of use:} Green Algorithms and ML CO2 Impact are highly user-friendly and require no programming skills, making them accessible to a broad audience. Energy Usage and Experiment Impact Tracker require more technical setup. Carbon Tracker simplifies usage via a web interface, whereas CodeCarbon and Cumulator require Python programming knowledge, targeting technically adept users.
    
    \item \textbf{Method of use:} Green Algorithms and ML CO2 Impact operate as web platforms accessible in browsers, requiring minimal setup. Energy Usage and Experiment Impact Tracker provide flexibility through desktop applications or code injection. Carbon Tracker, CodeCarbon, and Cumulator rely mainly on code injection, requiring integration into users’ coding workflows.
    
    \item \textbf{Functionality:} Green Algorithms, ML CO2 Impact, and Cumulator provide carbon footprint estimates, offering a general assessment. Energy Usage, Experiment Impact Tracker, and CodeCarbon deliver precise measurements suitable for detailed reporting. Carbon Tracker supports predictive analyses, aiding in forecasting and planning to reduce future emissions.
    
    \item \textbf{Measurement detail:} Green Algorithms and ML CO2 Impact provide global measurements, evaluating the overall footprint of the ML pipeline. In contrast, Energy Usage, Experiment Impact Tracker, Carbon Tracker, CodeCarbon, and Cumulator provide per-phase measurements, enabling users to identify and optimize specific stages such as preprocessing, training, evaluation, hyperparameter tuning, deployment, and inference.
\end{itemize}

Table~\ref{tab:carbon_footprint_comparison} presents a detailed comparison of the measurement tools with an in-depth overview of the features and capabilities of each tool.
\renewcommand{\arraystretch}{1.4}   
\setlength{\extrarowheight}{2pt}    

\begin{table*}[ht]
\caption{Comparison of the carbon footprint measurement tools}
\label{tab:carbon_footprint_comparison}
\small
\setlength{\tabcolsep}{6pt}
\renewcommand{\arraystretch}{1.4}
\setlength{\extrarowheight}{2pt}

\resizebox{\linewidth}{!}{%
\begin{tabular}{|p{1.2cm}|p{1.4cm}|p{1.37cm}|p{1.71cm}|p{1.72cm}|p{1.35cm}|p{1.71cm}|p{1.6cm}|}
\hline
\textbf{Criteria} & \textbf{Green Algorithms} & \textbf{ML CO2 Impact} & \textbf{Energy Usage} & \textbf{Experiment Impact Tracker} & \textbf{Carbon Tracker} & \textbf{Code-Carbon} & \textbf{Cumulator} \\ 
\hline
\textbf{C1} & Yes & Yes & Yes & Yes & Yes & Yes & Yes \\ 
\hline
\textbf{C2} & Limited & Limited & Limited & Limited & Abundant & Abundant & Abundant \\ 
\hline
\textbf{C3} & No coding required & No coding required & Requires technical setup & Requires technical setup & Web interface & Requires Python knowledge and environment & Requires Python knowledge and environment \\ 
\hline
\textbf{C4} & Web platform & Web platform & Desktop application or code injection & Desktop application or code injection & Code injection & Code injection & Code injection \\ 
\hline
\textbf{C5} & Estimation & Estimation & Measurement & Measurement & Prediction & Measurement & Estimation \\ 
\hline
\textbf{C6} & Global & Global & Per phase & Per phase & Per phase & Per phase & Per phase \\ 
\hline
\end{tabular}}

\end{table*}

To select the most suitable tool, we prioritize those offering clear and comprehensive documentation, ensuring that users with varying levels of technical expertise can accurately understand and utilize the tool. Next, we focus on tools capable of measuring the carbon footprint at each phase of the AI lifecycle, allowing users to evaluate energy consumption for specific stages, such as data preparation or model training. Finally, we emphasize tools that directly measure carbon emissions rather than providing estimates or predictions, as direct measurement delivers greater accuracy in environmental impact assessment.

Based on these criteria, CodeCarbon is identified as the preferred tool for quantifying carbon emissions. In the following section, we apply CodeCarbon to evaluate and compare the carbon footprints of six deep learning models.

\section{Computational Analysis of DL Models' Carbon Footprint}
\label{sec:Compt}

In this section, we evaluate and compare the carbon emissions of six DL models applied to a multi-label classification task, using CodeCarbon as the measurement tool.  

The primary objectives of this analysis are to:

\begin{itemize}
    \item Quantify the carbon footprint of six widely used DL models, both globally and for individual lifecycle phases, in order to determine the contribution of each phase to overall emissions.
    \item Compare global and per-phase emissions across the selected models to identify differences in environmental impact.
    \item Investigate the relationship between model accuracy and carbon footprint, highlighting potential trade-offs between performance and sustainability.
\end{itemize}
\subsection{Models}

We focus on widely adopted image-classification architectures. Table~\ref{tab:dl_models} summarizes the selected models considered in our empirical analysis, along with their key characteristics.

\begin{table}[!htbp]
    \centering
    \caption{Key characteristics of the deep learning models evaluated in the empirical study}
    \begin{tabular}{@{}lp{11cm}@{}}
    \toprule
        \textbf{Model}  & \textbf{Key characteristics} \\
    \midrule
        Simple CNN & Shallow stack of convolution and pooling layers with ReLU activations and dropout; fully connected classifier with softmax~\cite{cnn,10136725}. \\ \midrule
        U-Net  & Encoder–decoder with symmetric skip connections; designed for dense prediction and precise spatial localization~\cite{unet}. \\ \midrule
        ResNet  & Residual connections to address vanishing gradient problems in deep networks; achieves high accuracy with many layers~\cite{resnet}. \\ \midrule
        EfficientNet& Uses a compound scaling method that uniformly scales depth, width, and resolution based on a fixed set of scaling coefficients; achieves superior accuracy–efficiency trade-offs through mobile inverted bottleneck (MBConv) blocks and squeeze-and-excitation optimization~\cite{Saddami2024AdvancingGA}.\\ \midrule
        VGG16  & Very deep architecture with many small convolutional filters (3x3); successful early models, but can be computationally expensive~\cite{vgg16}. \\ \midrule

        VGG19  & Very deep architecture with many small convolutional filters (3x3); slightly more complex than VGG16 and computationally expensive~\cite{vgg19}. \\
    \bottomrule
    \end{tabular}
    \label{tab:dl_models}
\end{table}
\subsection{Work Environment}

Experiments were performed on a local machine equipped with an Intel(R) Core(TM) i5-10500 CPU running at 3.10 GHz and 16 GB of RAM. Each model was executed once under controlled and consistent system conditions to ensure fair and comparable results across all experiments.

Additionally, experiments were conducted within an Anaconda environment using Python 3.11.7 and CodeCarbon version 2.4.3rc1. 

CodeCarbon calculates CO$_2$ emissions using the formula: C=E×\(\mathrm{CI}\)~\cite{codecarbon}, where \(C\) represents the total carbon footprint (gCO$_2$e), \(E\) denotes the total energy consumed by the computational infrastructure (kWh), and \({CI}\) is the carbon intensity of the electricity used (gCO$_2$e/kWh), which varies by region. CodeCarbon accesses real-time databases to obtain \(\mathrm{CI}\) values corresponding to the location of execution. In our experiments, \(\mathrm{CI}\)=39.49 gCO$_2$e/kWh (ON, CAN).

Regions with higher carbon intensity \(\mathrm{CI}\) produce proportionally higher total emissions \(C\). Since \(C\) scales linearly with \(\mathrm{CI}\), relative comparisons across models and experimental settings remain valid under uniform shifts in regional carbon intensity.

\subsection{Data Exploration}

In the development of our DL models, we utilized the CIFAR-10 dataset. The dataset and its preprocessing are described in detail below.

\subsubsection{Dataset Description}

The CIFAR-10 dataset is a subset of the Tiny Images dataset~\cite{4531741} and comprises 60,000 color images of size 32×32 pixels. These images are evenly distributed across 10 object classes: Airplane, Automobile, Bird, Cat, Deer, Dog, Frog, Horse, Ship, and Truck~\cite{Krizhevsky2009LearningML}. For our experiments, 50,000 images were used for training, while the remaining 10,000 images were reserved for testing.

\subsubsection{Data Preparation}

The following steps were applied to prepare the CIFAR-10 dataset for input into the selected deep learning models:

\begin{itemize}
    \item \textbf{Data Quality:} To improve the training set quality, we implemented a function to detect blurry images. Only 0.02\% of images were found to be blurry and were removed to prevent potential degradation of model performance. Additionally, we identified and excluded dark images from the dataset, resulting in a cleaned and higher-quality training set.
    
    \item \textbf{Reshaping and Normalization:} All images were resized to a consistent shape and normalized to the range [0, 1]. Pixel values were converted to \texttt{float32} and divided by 255, ensuring uniform input across the dataset.
    
    \item \textbf{One-Hot Encoding:} Class labels were transformed into one-hot vectors using TensorFlow’s \texttt{to\_categorical} function. This encoding allows categorical classification, representing each class label as a binary vector with a single high (1) value indicating the correct class.
    
    \item \textbf{Data Augmentation:} To enhance dataset diversity and increase the effective training size, we applied a series of image transformations, including rotations, flips, and shifts, helping the models generalize better to unseen data.
\end{itemize}

\section{Results}
\label{sec:results}
We trained the six models in the same computational environment to maintain consistent operating conditions.

\begin{figure}[!htpb]
\centering
\includegraphics[width=1\linewidth]{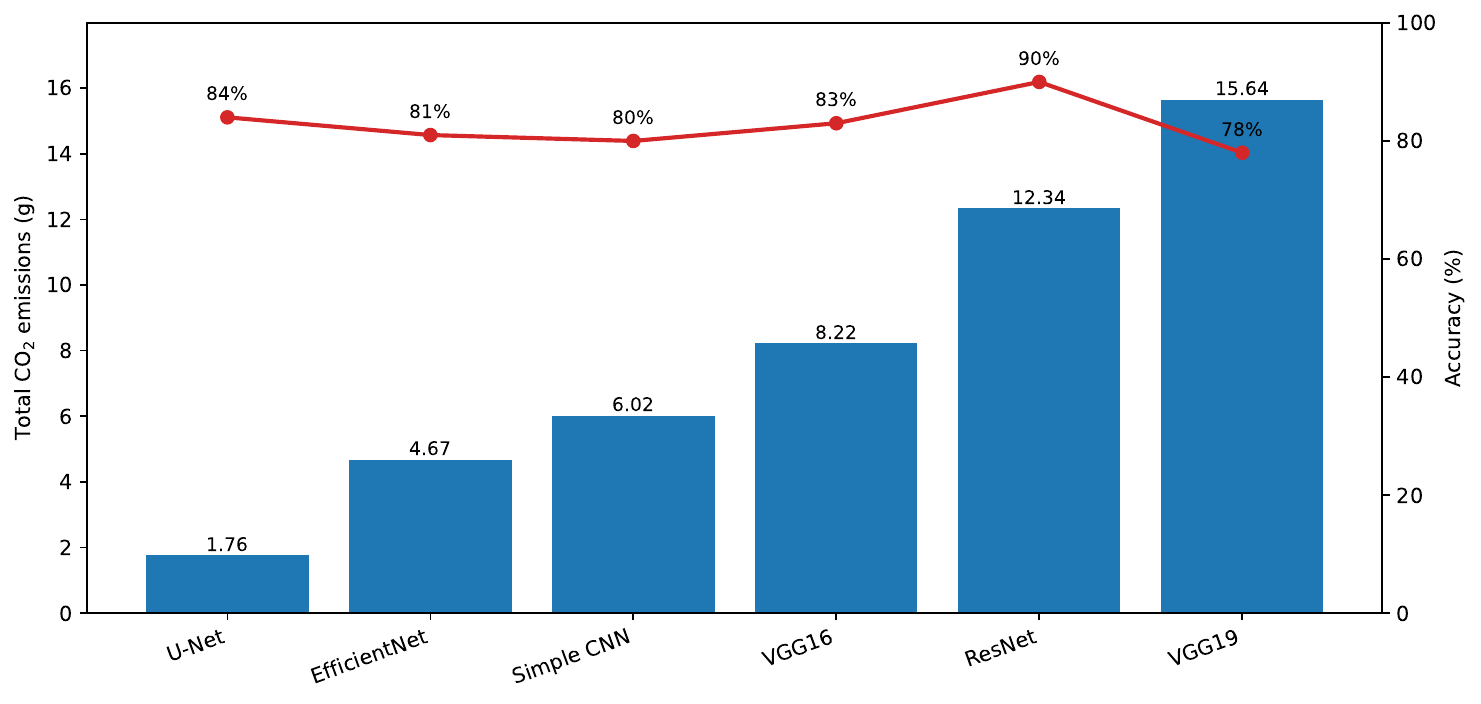}
\caption{Total carbon emissions and classification accuracy for the six evaluated models}
\label{fig:total}
\end{figure}
We compared the accuracy and carbon emissions of the selected models. Fig.~\ref{fig:total} illustrates the total carbon emissions (in grams) and accuracies associated with each model.

Among the six models, UNet and EfficientNet exhibit the lowest total emissions, approximately 1.76 g and 4.67 g, while maintaining solid accuracies of 84\% and 81\%, respectively.
The Simple CNN also demonstrates relatively low emissions ($\approx$ 6.02 g) and competitive accuracy of 80\%.
By contrast, VGG16 and ResNet consume more energy ($\approx$ 8.22 g and $\approx$ 12.34 g, respectively) but achieve higher accuracies of 83\% and 90\%.
Finally, VGG19 records both the highest emissions ($\approx$ 15.64 g) and the lowest accuracy (78\%), indicating low efficiency.

Overall, UNet and EfficientNet emerge as the most environmentally efficient models, offering a favorable balance between performance and carbon footprint. ResNet, on the other hand, achieves the best accuracy but at the cost of significantly higher emissions.

\begin{figure}[!h]
    \centering
    \includegraphics[width=0.8\linewidth]{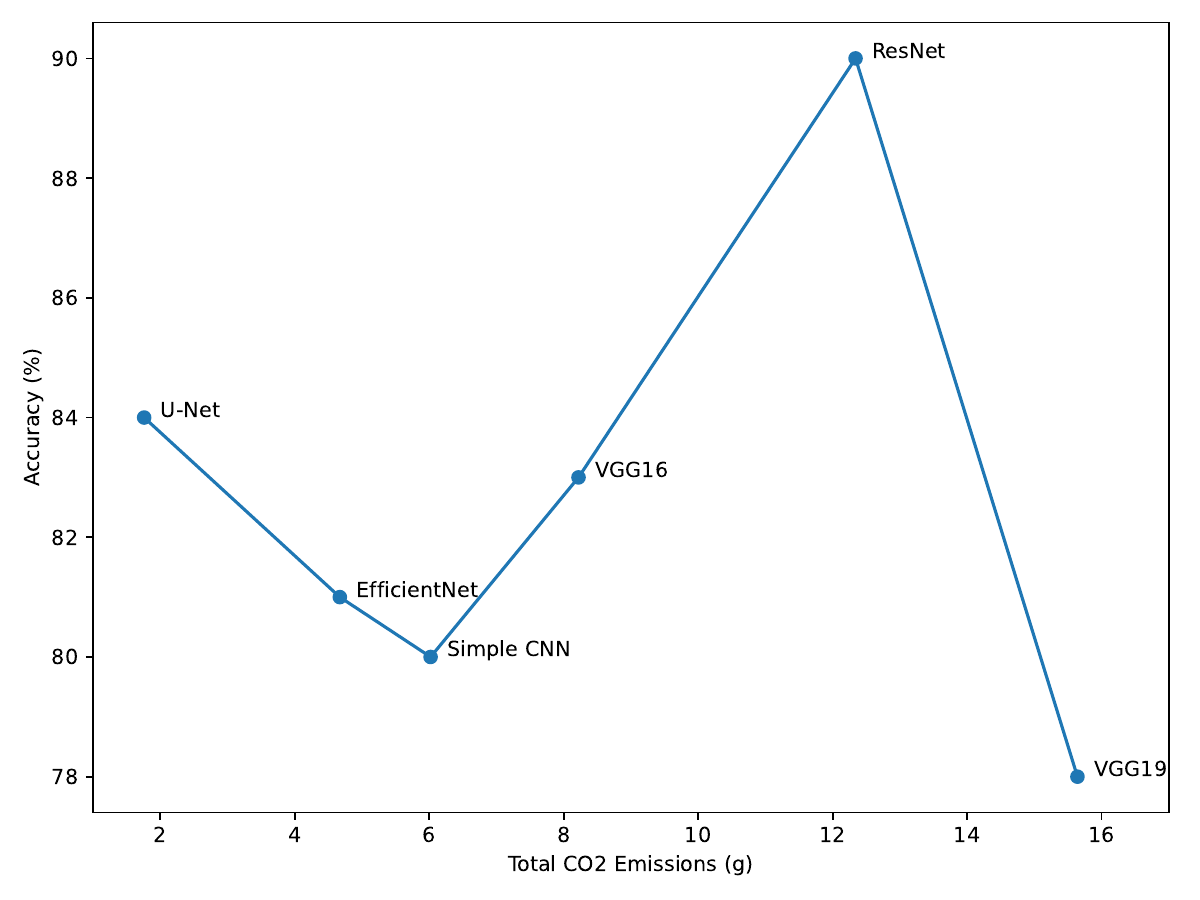}
    \caption{Trade-off between model accuracy and total carbon emissions for the six evaluated models}
    \label{fig:carbon_accuracy_tradeoff}
\end{figure}
  
 Additionally, Fig.\ref{fig:carbon_accuracy_tradeoff} shows that increasing carbon emissions does not lead to a monotonic improvement in accuracy. Although accuracy generally increases with higher emissions up to a point, the trend eventually saturates and then declines for the highest-emission model. In particular, some low-emission models achieve competitive accuracy, whereas the model with the highest emissions does not. \\
 
To quantify the environmental cost of predictive performance, we define the Emissions per Accuracy Point (EAP)~\cite{strubell2019energy,henderson2020towards,patterson2021carbon} as the grams of CO$_2$e emitted per accuracy point. For a given model $i$, EAP is defined as:
\begin{equation}
EAP_i = \frac{C_i}{A_i}
\end{equation}

Where $C_i$ is the total carbon emissions (in gCO$_2$e) of model $i$, and $A_i$ is its accuracy level. Lower EAP values indicate better carbon efficiency.

To enable direct comparison across models, we further compute a min--max normalized EAP score:

\begin{equation}
EAP^{Norm}_i = \frac{EAP_i - \min(EAP)}{\max(EAP) - \min(EAP)}
\end{equation}

which rescales efficiency to the range $[0,1]$, where 0 represents the most carbon-efficient model.

\begin{table}[!htpb]
\centering
\caption{Emissions per Accuracy Point (EAP) for the six evaluated models}
\label{tab:eap}
\begin{tabular}{lcccc}
\hline
\textbf{Model} & \textbf{ Carbon Emissions (g)} & \textbf{Accuracy} & \textbf{$EAP$} & \textbf{$EAP^{Norm}_i$} \\
\hline

U-Net        & 1.76  & 0.84 & 2.10 & 0.00 \\
EfficientNet & 4.67  & 0.81 & 5.77 & 0.22 \\
Simple CNN   & 6.02  & 0.80 & 7.53 & 0.31 \\
VGG16        & 8.22  & 0.83 & 9.91 & 0.44 \\
ResNet       & 12.34 & 0.90 & 13.71 & 0.66 \\
VGG19        & 15.64 & 0.78 & 20.05 & 1.00 \\
\hline
\end{tabular}
\end{table}

The results in Table~\ref{tab:eap} show that U-Net achieves the lowest EAP (2.10 gCO$_2$e per accuracy point) and the lowest normalized efficiency score (0.00), indicating the most favorable sustainability profile. EfficientNet (0.22) and Simple CNN (0.31) constitute an intermediate-efficiency tier, reflecting a reasonable trade-off between carbon emissions and predictive performance. VGG16 (0.44) and ResNet (0.66) exhibit substantially higher carbon costs per unit of accuracy. Although ResNet achieves the highest absolute accuracy (0.90), its normalized efficiency indicates that these performance gains come at a disproportionately higher environmental cost. VGG19 records the highest normalized score (1.00), representing the least efficient model, as it combines the highest emissions with lower accuracy.

\begin{figure}[!htpb]
\centering
\includegraphics[width=0.9\linewidth]{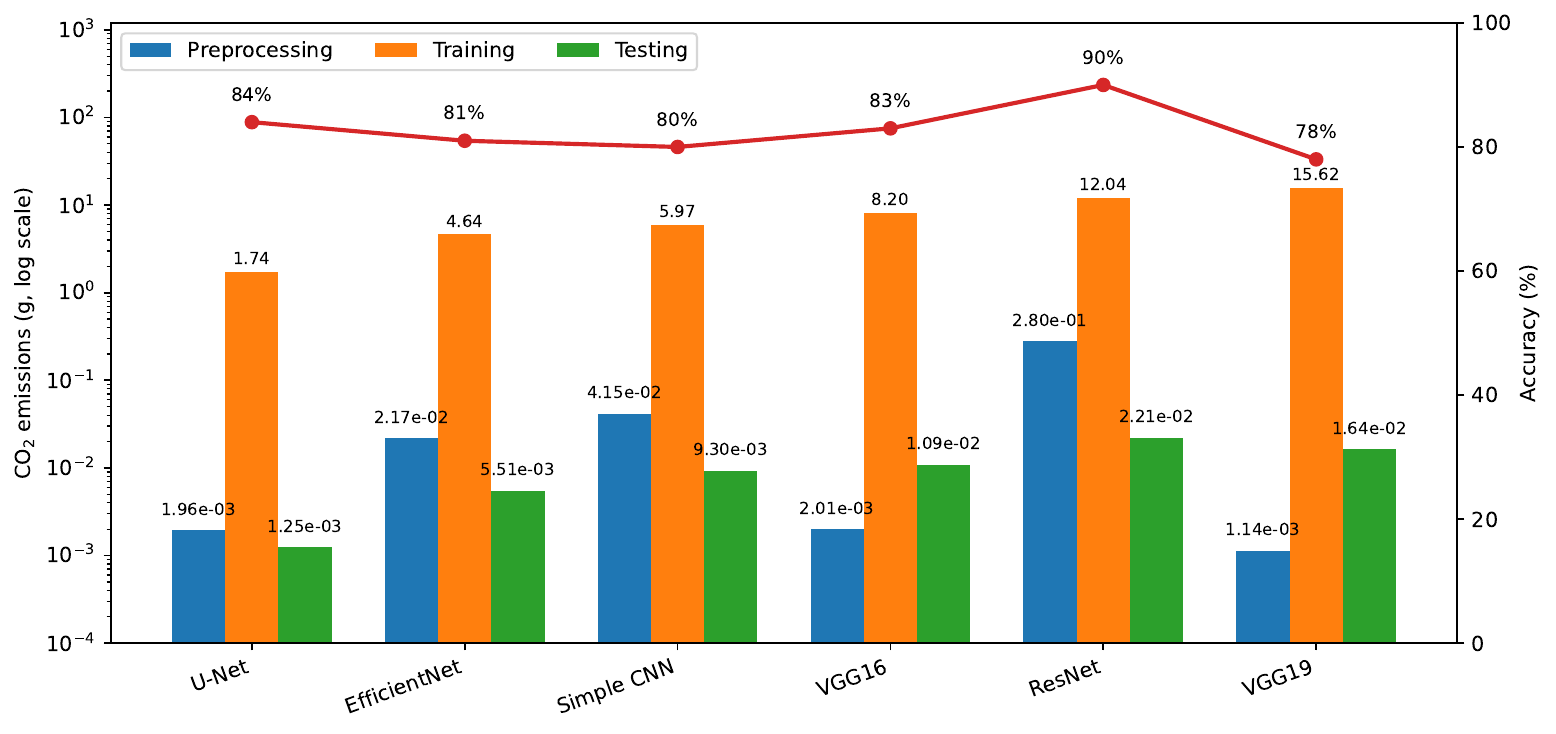}
\caption{Per-phase carbon emissions and accuracy for each model (logarithmic scale)}
\label{fig:phase}
\end{figure}

For further investigation, we measured carbon emissions by phase across the model's lifecycle to identify which stages contribute most significantly to the overall carbon footprint. Fig.~\ref{fig:phase} illustrates the breakdown of total emissions into the three main computational phases—preprocessing, training, and testing—plotted on a logarithmic scale to highlight the differences in magnitude. 

Across all models, the training phase overwhelmingly dominates the total carbon footprint, contributing orders of magnitude more emissions than preprocessing and testing. 
The preprocessing and testing phases exhibit negligible emissions (on the order of $10^{-3}$–$10^{-1}$~g), reinforcing that the majority of the environmental impact arises during model optimization.

Complex architectures such as ResNet, VGG16, and VGG19 exhibit significantly higher emissions during the training phase—up to two orders of magnitude higher than lighter models such as UNet, Simple CNN, and EfficientNet. 
This trend highlights a clear trade-off between model complexity and environmental cost: deeper, parameter-heavy networks achieve higher accuracy but at the expense of increased carbon emissions.

Overall, these results emphasize that optimization efforts should primarily target the training phase, where the largest reductions in energy use and emissions can be achieved, while preprocessing and testing offer limited potential for further improvement.

\section{Discussion}
\label{sec:discussion}

The results were obtained from a single execution in a CPU-based experimental environment, using a carbon intensity representative of the Ontario (Canada) electricity grid. The observed differences in carbon emissions are mainly due to the architectural complexity and computational demands of the evaluated models. Across all selected models, the training phase remains the dominant contributor to emissions, whereas preprocessing and testing contribute negligibly. This finding underscores the need to focus optimization efforts on the training stage, where even modest efficiency gains can yield significant environmental benefits.

Beyond absolute emission values, the EAP analysis provides a more nuanced interpretation of sustainability. While raw emissions quantify environmental impact, EAP captures the carbon cost required to obtain one unit of predictive accuracy. The results show substantial variation in carbon efficiency across architectures. U-Net achieves the lowest EAP, indicating the most favorable balance between performance and emissions. In contrast, VGG19 records the highest EAP, reflecting a disproportionately high environmental cost relative to its predictive performance. The normalized EAP scores further confirm this ranking and reveal clear efficiency tiers among the evaluated models.

Importantly, the results reveal no direct positive correlation between model accuracy and carbon emissions. For instance, ResNet achieves the highest accuracy among all models, yet its EAP value indicates a significantly higher carbon cost per unit of accuracy compared to more efficient architectures such as U-Net. Conversely, VGG19 combines high emissions with lower accuracy, resulting in the least favorable efficiency profile. These findings demonstrate that architectural complexity and computational intensity do not necessarily translate into proportionally improved predictive performance when environmental cost is considered.

The trade-off between accuracy and emissions becomes particularly relevant in real-world contexts. In high-stakes domains such as healthcare or autonomous systems, maximizing accuracy remains essential, as even marginal improvements can be critical. However, the EAP analysis highlights that such gains may come at increasing environmental cost. Conversely, in large-scale or resource-intensive applications—such as climate modeling, sustainability analytics, or edge AI deployments—carbon efficiency may take precedence. In these scenarios, selecting architectures with lower EAP values can substantially reduce environmental impact while maintaining competitive predictive performance.

Ultimately, these findings support a domain-sensitive approach to AI development: practitioners should balance accuracy and carbon efficiency in light of the application’s societal and ecological context. By incorporating normalized carbon efficiency metrics such as EAP into model evaluation, the AI community can move toward genuinely sustainable AI practices in which technological advancement and environmental responsibility evolve in parallel.

\section{Threats to Validity}
\label{sec:threatsValidity}

\textbf{Construct Validity} \\
In this work, carbon emissions were estimated using the CodeCarbon tool, which computes emissions based on the regional energy carbon intensity of Ontario, Canada (where we performed our analysis), hardware utilization, and execution time. Although CodeCarbon is widely adopted and validated in prior Green AI research, the measurements may be slightly affected by system-level variability. To mitigate this threat, all experiments were conducted under consistent hardware and software configurations, with identical execution settings and controlled experimental conditions to ensure fair and reliable comparisons between models. Furthermore, model performance was evaluated using standard metrics, ensuring an appropriate and accurate representation of model effectiveness.\\

\noindent\textbf{Internal Validity} \\
The benchmarked models used in our experimental comparison are primarily convolutional neural networks. While these models are widely used in computer vision, including more complex architectures may present different emission–performance characteristics. Investigating these models will be necessary to broaden the relevance of the findings. 
Additionally, carbon footprint measurements were obtained from single experimental runs. Although CodeCarbon provides reliable runtime measurements, energy consumption may vary slightly due to system-level variability. To mitigate this risk, unnecessary background processes were minimized, and identical execution conditions were maintained for all models. Future work will include repeated experiments and statistical analysis to further improve measurement robustness and confidence in quantitative comparisons.\\

\noindent\textbf{External Validity}\\
Our experiments were conducted on a single dataset and a CPU-based environment. While this controlled setup enables reproducible measurements of carbon emissions, the results may not generalize to GPU- or cloud-based infrastructures, where different power profiles, parallelization characteristics, and data center energy mixes may alter the observed carbon–performance trade-offs. Extending the framework to diverse hardware configurations and regional energy profiles represents an important direction for future work.

\section{Conclusion and Future Directions}
\label{sec:conc}

This work provided a comprehensive investigation into the environmental sustainability of AI, bridging a systematic literature review with empirical experimentation. Through a rigorous SLR and mapping process based on the PRISMA framework, we identified and analyzed the existing literature on Green AI and Green DL. Furthermore, we conducted a structured comparison of existing carbon footprint measurement tools, highlighting differences in transparency, granularity, hardware compatibility, and regional carbon intensity modeling. Our review revealed that, although Green AI research is rapidly expanding, it still lacks standardized methodologies for measuring and reporting the energy consumption and carbon emissions of deep learning models. Motivated by this gap, the study presented an empirical analysis of the relationship between model accuracy and carbon emissions across several convolutional neural network architectures, including UNet, Simple CNN, VGG16, ResNet, VGG19, and EfficientNet. 

By maintaining a consistent computational environment, we quantified the energy consumption and corresponding carbon emissions for each model using the CodeCarbon tool, revealing that the training phase accounts for the overwhelming majority of the overall environmental footprint within the evaluated CPU-based experimental setting. In contrast, the preprocessing and testing phases contribute negligibly to total emissions.

Additionally, the results indicate that higher architectural complexity does not necessarily yield proportionally higher accuracy. For example, ResNet achieved the highest accuracy without the highest emissions, whereas VGG19 exhibited the highest emissions and the lowest accuracy. 
These findings underscore that more powerful models are not always more efficient or sustainable, reinforcing the need to evaluate both performance and environmental cost during model selection.

From a sustainability perspective, optimization efforts should prioritize reducing the energy consumption of the training phase through methods such as model pruning, quantization, efficient hyperparameter tuning, and scheduling training on renewable-powered infrastructure. 
Moreover, the adoption of carbon-aware benchmarking—where emissions are reported alongside traditional performance metrics—can help drive a culture of environmentally responsible AI development.

It is important to note that these empirical findings are derived from a single run within a CPU-based experimental environment and are therefore most directly applicable to small-to-moderate scale training scenarios or resource-constrained deployments. In GPU-based or distributed training environments, the carbon–performance trade-off may differ qualitatively from the patterns observed in this study. GPUs typically provide higher computational throughput, which can shorten training time but increase instantaneous power demand, while distributed training introduces communication and infrastructure overheads. As a result, total emissions may vary depending on hardware efficiency, utilization, and data center energy mix.

Looking ahead, our research will extend beyond carbon footprint measurement toward the development of greener AI algorithms and the optimization of the full machine learning pipeline. As a next step, we plan to experiment with the optimization techniques identified in our review tables (\ref{tab:prep}, \ref{tab:train} and \ref{tab:inf}), including pruning, quantization, and mixed-precision training, to evaluate their real-world applicability. This will allow us to determine whether these approaches effectively reduce emissions while maintaining model performance and whether they represent viable strategies for making existing models more environmentally friendly.

In addition, future work will investigate the impact of computational infrastructure diversity on AI sustainability. In particular, we plan to evaluate how alternative hardware backends—including GPUs and cloud-based environments—affect training efficiency, energy consumption, and carbon intensity. Such investigations will help generalize our findings across heterogeneous deployment environments and provide a more comprehensive understanding of infrastructure-aware sustainability trade-offs.

Another important direction is the evaluation of large-scale foundation models, which are rapidly becoming central to modern AI systems and may introduce substantially higher environmental costs due to their scale and training requirements. Extending carbon footprint analysis to these architectures will improve understanding of the sustainability implications of emerging large-capacity models and support the development of responsible deployment strategies.

Beyond centralized training scenarios, another promising direction for future work is the study of energy efficiency in federated and distributed learning settings. In such scenarios, the environmental impact is influenced not only by local computation but also by communication overhead, repeated local updates, and system heterogeneity, which introduce new sustainability trade-offs. In parallel, we plan to further investigate the deployment of AI workloads on renewable-powered or carbon-aware infrastructures, in which training and inference can be scheduled based on the availability of low-carbon energy. These system-level and deployment-oriented strategies could complement algorithmic optimizations and further reduce the environmental footprint of AI systems in real-world settings. Ultimately, the goal is to contribute to a more comprehensive understanding of how to jointly optimize AI performance and ecological impact, paving the way for truly sustainable AI systems.

\section*{Acknowledgment}
Samar Garrab reports that financial support was provided by the Canadian Defence Academy Research Program and the Mitacs Globalink Research Internship.
\section*{Published Article Notice}
This manuscript is the accepted author version of the following published article:\\
Garrab, S., Boughriou, S. \& BenSassi, M. Towards sustainable artificial intelligence: a comprehensive review and comparative analysis of deep learning models’ carbon footprint. Appl Intelligence 56, 173 (2026). \\
The final authenticated publication is available at:\\
https://doi.org/10.1007/s10489-026-07208-y\\
Please cite the published version.

\section*{Authors Contribution}%
\textbf{Samar Garrab:} Conceptualization, Writing - Review \& editing, Validation, Supervision, Project administration, Methodology, and Funding acquisition.

\noindent\textbf{Sarra Boughriou:} Writing - Original draft \& editing, Methodology, Investigation, and Data curation. 

\noindent\textbf{Manel Ben Sassi:} Review, Validation, and Supervision.

\section*{Data Availability}
The dataset used and analyzed during the current study is available in Torralba et al.~\cite{4531741} article. Furthermore, you can find our experiments online at \url{https://figshare.com/s/5fd9622e89ad5c525ccd}. 

\section*{Conflict of Interest}%
 The authors declare that they have no known conflict of interest regarding the publication of this manuscript.
 \newpage

\bibliography{referenceRevised2}

@article{anthony2020carbontracker,
  title={Carbontracker: Tracking and predicting the carbon footprint of training deep learning models},
  author={Anthony, Lasse F Wolff and Kanding, Benjamin and Selvan, Raghavendra},
  journal={arXiv preprint},
  year={2020.},
doi={https://doi.org/10.48550/arXiv.2007.03051}

}

@article{ba2016layer,
  title={Layer normalization},
  author={Ba, Jimmy Lei and Kiros, Jamie Ryan and Hinton, Geoffrey E},
  journal={arXiv preprint},
  year={2016.},
doi={https://doi.org/10.48550/arXiv.1607.06450}
}

@article{ba2014deep,
  title={Do deep nets really need to be deep?},
  author={Ba, Jimmy and Caruana, Rich},
  journal={Advances in neural information processing systems},
  volume={27},
  year={2014},
doi={https://doi.org/10.48550/arXiv.1312.6184}
}

@inproceedings{bannour2021evaluating,
  title={Evaluating the carbon footprint of NLP methods: a survey and analysis of existing tools},
  author={Bannour, Nesrine and Ghannay, Sahar and N{\'e}v{\'e}ol, Aur{\'e}lie and Ligozat, Anne-Laure},
  booktitle={Proceedings of the second workshop on simple and efficient natural language processing},
  pages={11--21},
  year={2021.},
  doi = {10.18653/v1/2021.sustainlp-1.2}
}

@inproceedings{belilovsky2019greedy,
  title={Greedy layerwise learning can scale to imagenet},
  author={Belilovsky, Eugene and Eickenberg, Michael and Oyallon, Edouard},
  booktitle={International conference on machine learning},
  pages={583--593},
  year={2019},
  organization={PMLR},
doi={https://doi.org/10.48550/arXiv.1812.11446}
}

@misc{dhar2020carbon,
  title={The carbon impact of artificial intelligence},
  author={Dhar, Payal},
  year={2020},
  publisher={Nature Publishing Group UK London.},
doi={https://doi.org/10.1038/s42256-020-0219-9}
}

@inproceedings{gutierrez2022analysing,
  title={Analysing the energy impact of different optimisations for machine learning models},
  author={Guti{\'e}rrez, Mar{\'\i}a and Moraga, Ma {\'A}ngeles and Garc{\'\i}a, F{\'e}lix},
  booktitle={2022 international conference on ICT for sustainability (ICT4S)},
  pages={46--52},
  year={2022},
  organization={IEEE.},
  doi={10.1109/ICT4S55073.2022.00016}
}

@article{henderson2020towards,
  title={Towards the systematic reporting of the energy and carbon footprints of machine learning},
  author={Henderson, Peter and Hu, Jieru and Romoff, Joshua and Brunskill, Emma and Jurafsky, Dan and Pineau, Joelle},
  journal={Journal of Machine Learning Research},
  volume={21},
  number={248},
  pages={1--43},
  year={2020.},
doi={https://doi.org/10.48550/arXiv.2002.05651}
}

@article{hinton2015distilling,
  title={Distilling the knowledge in a neural network},
  author={Hinton, Geoffrey E and Vinyals, Oriol and Dean, Jeff},
  journal={arXiv preprint},
pages= {1-9},
  year={2015.},
doi={https://doi.org/10.48550/arXiv.1503.02531}
}

@article{ioffe2015batch,
  title={Batch normalization: Accelerating deep network training by reducing internal covariate shift},
  author={Ioffe, Sergey},
  journal={arXiv preprint},
  year={2015.}, 
doi={https://doi.org/10.48550/arXiv.1502.03167}
}

@article{lacoste2019quantifying,
  title={Quantifying the carbon emissions of machine learning},
  author={Lacoste, Alexandre and Luccioni, Alexandra and Schmidt, Victor and Dandres, Thomas},
  journal={arXiv preprint},
  year={2019.},
doi={
https://doi.org/10.48550/arXiv.1910.09700}
}

@article{lannelongue2021green,
  title={Green algorithms: quantifying the carbon footprint of computation},
  author={Lannelongue, Lo{\"\i}c and Grealey, Jason and Inouye, Michael},
  journal={Advanced science},
  volume={8},
  number={12},
  pages={2100707},
  year={2021},
  publisher={Wiley Online Library.},
doi={https://doi.org/10.48550/arXiv.2007.07610}
}

@article{lecun1990optimal,
  title={Optimal brain damage},
  author={LeCun, Yann and Denker, John and Solla, Sara},
  journal={Advances in neural information processing systems},
  volume={2},
  year={1989},
url={https://api.semanticscholar.org/CorpusID:7785881}
}

@inproceedings{li2016evaluating,
  title={Evaluating the energy efficiency of deep convolutional neural networks on CPUs and GPUs},
  author={Li, Da and Chen, Xinbo and Becchi, Michela and Zong, Ziliang},
  booktitle={2016 IEEE international conferences on big data and cloud computing (BDCloud), social computing and networking (SocialCom), sustainable computing and communications (SustainCom)(BDCloud-SocialCom-SustainCom)},
  pages={477--484},
  year={2016},
  organization={IEEE},
 doi={10.1109/BDCloud-SocialCom-SustainCom.2016.76}
}

@article{lottick2019energy,
  title={Energy Usage Reports: Environmental awareness as part of algorithmic accountability},
  author={Lottick, Kadan and Susai, Silvia and Friedler, Sorelle A and Wilson, Jonathan P},
  journal={arXiv preprint},
  year={2019.}, 
doi={
https://doi.org/10.48550/arXiv.1911.08354}
}

@article{menghani2023efficient,
  title={Efficient deep learning: A survey on making deep learning models smaller, faster, and better},
  author={Menghani, Gaurav},
  journal={ACM Computing Surveys},
  volume={55},
  number={12},
  pages={1--37},
  year={2023},
  publisher={ACM New York, NY.}
}

@article{mishkin2015all,
  title={All you need is a good init},
  author={Mishkin, Dmytro and Matas, Jiri},
  journal={arXiv preprint},
  year={2015.}, 
doi={https://doi.org/10.48550/arXiv.1511.06422}
}

@article{natarajan2022theoretical,
  title={A theoretical review on AI affordances for sustainability},
  author={Natarajan, Harish Karthi and de Paula, Danielly and Dremel, Christian and Uebernickel, Falk},
  journal={AMCIS 2022 Proceedings},
  year={2022},
  publisher={Association for Information Systems (AIS)},
url={https://aisel.aisnet.org/amcis2022/sig_green/sig_green/13}
}

@inproceedings{rafiullah2023ai,
  title={AI and energy efficiency},
  author={Omar, Rafiullah},
  booktitle={2023 IEEE 20th International Conference on Software Architecture Companion (ICSA-C)},
  pages={141--144},
  year={2023},
  organization={IEEE},
  doi={10.1109/ICSA-C57050.2023.00040}}

@article{raghuraman2018quantizing,
  title={Quantizing deep convolutional networks for efficient inference: A whitepaper},
  author={Krishnamoorthi, Raghuraman},
  journal={arXiv preprint},
  year={2018.}, 
doi={https://doi.org/10.48550/arXiv.1806.08342}
}

@article{rolnick2019tackling,
  title={Tackling climate change with machine learning},
  author={Rolnick, David and Donti, Priya L and Kaack, Lynn H and Kochanski, Kelly and Lacoste, Alexandre and Sankaran, Kris and Ross, Andrew Slavin and Milojevic-Dupont, Nikola and Jaques, Natasha and Waldman-Brown, Anna and others},
  journal={ACM Computing Surveys (CSUR)},
  volume={55},
  number={2},
  pages={1--96},
  year={2022},
  publisher={ACM New York, NY.}, 
doi={
https://doi.org/10.48550/arXiv.1906.05433
}
}

@article{schwartz2020green,
  title={Green ai},
  author={Schwartz, Roy and Dodge, Jesse and Smith, Noah A and Etzioni, Oren},
  journal={Communications of the ACM},
  volume={63},
  number={12},
  pages={54--63},
  year={2020},
  publisher={ACM New York, NY, USA}, 
doi = {10.1145/3381831},
}

@inproceedings{strubell2019energy,
  title={Energy and policy considerations for deep learning in NLP},
  author={Strubell, Emma and Ganesh, Ananya and McCallum, Andrew},
  booktitle={Proceedings of the 57th annual meeting of the association for computational linguistics},
  pages={3645--3650},
  year={2019.}, 
doi={
https://doi.org/10.48550/arXiv.1906.02243
}
}

@article{trebaol2020cumulator,
  title={CUMULATOR—a tool to quantify and report the carbon footprint of machine learning computations and communication in academia and healthcare},
  author={Tr{\'e}baol, Tristan},
  journal={Semester Project, EPFL},
  year={2020},
url={https://api.semanticscholar.org/CorpusID:235654375}
}

@article{verdecchia2023systematic,
  title={A systematic review of Green AI},
  author={Verdecchia, Roberto and Sallou, June and Cruz, Lu{\'\i}s},
  journal={Wiley Interdisciplinary Reviews: Data Mining and Knowledge Discovery},
  volume={13},
  number={4},
  pages={e1507},
  year={2023},
  publisher={Wiley Online Library.},
  doi={https://doi.org/10.1002/widm.1507}
}

@article{whang2023data,
  title={Data collection and quality challenges in deep learning: A data-centric ai perspective},
  author={Whang, Steven Euijong and Roh, Yuji and Song, Hwanjun and Lee, Jae-Gil},
  journal={The VLDB Journal},
  volume={32},
  number={4},
  pages={791--813},
  year={2023},
  publisher={Springer.}, 
doi={
https://doi.org/10.48550/arXiv.2112.06409}
}

@article{wu2022sustainable,
  title={Sustainable ai: Environmental implications, challenges and opportunities},
  author={Wu, Carole-Jean and Raghavendra, Ramya and Gupta, Udit and Acun, Bilge and Ardalani, Newsha and Maeng, Kiwan and Chang, Gloria and Aga, Fiona and Huang, Jinshi and Bai, Charles and others},
  journal={Proceedings of machine learning and systems},
  volume={4},
  pages={795--813},
  year={2022.}, 
doi={
https://doi.org/10.48550/arXiv.2111.00364}
}

@inproceedings{wu2018group,
  title={Group normalization},
  author={Wu, Yuxin and He, Kaiming},
  booktitle={Proceedings of the European conference on computer vision (ECCV)},
  pages={3--19},
  year={2018.}, 
doi={
https://doi.org/10.48550/arXiv.1803.08494}
}

@article{wynsberghe2021sustainable,
  title={Sustainable AI: AI for sustainability and the sustainability of AI},
  author={Van Wynsberghe, Aimee},
  journal={AI and Ethics},
  volume={1},
  number={3},
  pages={213--218},
  year={2021},
  publisher={Springer.}, 
doi={https://doi.org/10.1007/s43681-021-00043-6}
}

@article{xu2021survey,
  title={A survey on green deep learning},
  author={Xu, Jingjing and Zhou, Wangchunshu and Fu, Zhiyi and Zhou, Hao and Li, Lei},
  journal={arXiv preprint},
  year={2021.}, 
doi={
https://doi.org/10.48550/arXiv.2111.05193
}
}

@misc{
yu2018fast,
title={Fast and Accurate Text Classification: Skimming, Rereading and Early Stopping},
author={Keyi Yu and Yang Liu and Alexander G. Schwing and Jian Peng},
year={2018},
url={https://openreview.net/forum?id=ryZ8sz-Ab},
}

@article{zhang2019fixup,
  title={Fixup initialization: Residual learning without normalization},
  author={Zhang, Hongyi and Dauphin, Yann N and Ma, Tengyu},
  journal={arXiv preprint},
  year={2019.},
doi={https://doi.org/10.48550/arXiv.1901.09321}
}

@inproceedings{budennyy2022eco2AI,
  title={Eco2ai: carbon emissions tracking of machine learning models as the first step towards sustainable ai},
  author={Budennyy, Semen Andreevich and Lazarev, Vladimir Dmitrievich and Zakharenko, Nikita Nikolaevich and Korovin, Aleksei N and Plosskaya, OA and Dimitrov, Denis Valer’evich and Akhripkin, VS and Pavlov, IV and Oseledets, Ivan Valer’evich and Barsola, Ivan Segundovich and others},
  booktitle={Doklady mathematics},
  volume={106},
  number={Suppl 1},
  pages={S118--S128},
  year={2022},
  organization={Springer.}, 
doi={https://doi.org/10.1134/S1064562422060230}
}

@article{pachot2022towards,
  title={Towards sustainable artificial intelligence: an overview of environmental protection uses and issues},
  author={Pachot, Arnault and Patissier, C{\'e}line},
  journal={arXiv preprint},
  year={2022.},
doi={https://doi.org/10.48550/arXiv.2212.11738}
}

@article{zhou2023opportunities,
  title={On the opportunities of green computing: A survey},
  author={Zhou, You and Lin, Xiujing and Zhang, Xiang and Wang, Maolin and Jiang, Gangwei and Lu, Huakang and Wu, Yupeng and Zhang, Kai and Yang, Zhe and Wang, Kehang and others},
  journal={arXiv preprint},
  year={2023.},
doi={https://doi.org/10.48550/arXiv.2311.00447}
}

@misc{codecarbon,
  title={CodeCarbon: Track and reduce your carbon emissions in machine learning},
  author={{ML CO2}},
  year={2023},
  url={https://github.com/mlco2/codecarbon},
  note={Accessed: 2024-06-26}
}

@article{barbierato2024towards,
  title={Toward green ai: A methodological survey of the scientific literature},
  author={Barbierato, Enrico and Gatti, Alice},
  journal={IEEE Access},
  volume={12},
  pages={23989--24013},
  year={2024},
  publisher={IEEE},
  doi={10.1109/ACCESS.2024.3360705}}

@article{Krizhevsky2009LearningML,
  title={Learning multiple layers of features from tiny images},
  author={Krizhevsky, Alex and Hinton, Geoffrey and others},
  year={2009},
volume={18268744},
journal={University of Toronto},
  publisher={Toronto, ON, Canada}
}

@article{4531741,
  title={80 million tiny images: A large data set for nonparametric object and scene recognition},
  author={Torralba, Antonio and Fergus, Rob and Freeman, William T},
  journal={IEEE transactions on pattern analysis and machine intelligence},
  volume={30},
  number={11},
  pages={1958--1970},
  year={2008},
  publisher={IEEE},
  doi={10.1109/TPAMI.2008.128}}

@inproceedings{snowballing,
  title={Guidelines for snowballing in systematic literature studies and a replication in software engineering},
  author={Wohlin, Claes},
  booktitle={Proceedings of the 18th international conference on evaluation and assessment in software engineering},
  pages={1--10},
  year={2014},
doi = {10.1145/2601248.2601268}
}

@inproceedings{unet,
  title={U-net: Convolutional networks for biomedical image segmentation},
  author={Ronneberger, Olaf and Fischer, Philipp and Brox, Thomas},
  booktitle={International Conference on Medical image computing and computer-assisted intervention},
  pages={234--241},
  year={2015},
  organization={Springer},
doi={
 https://doi.org/10.1007/978-3-319-24574-4_28
}
}

@inproceedings{resnet,
  title={Deep residual learning for image recognition},
  author={He, Kaiming and Zhang, Xiangyu and Ren, Shaoqing and Sun, Jian},
  booktitle={Proceedings of the IEEE conference on computer vision and pattern recognition},
  pages={770--778},
  year={2016},
url={https://openaccess.thecvf.com/content_cvpr_2016/html/He_Deep_Residual_Learning_CVPR_2016_paper.html}
}

@inproceedings{vgg16,
  title={Compressed residual-VGG16 CNN model for big data places image recognition},
  author={Qassim, Hussam and Verma, Abhishek and Feinzimer, David},
  booktitle={2018 IEEE 8th annual computing and communication workshop and conference (CCWC)},
  pages={169--175},
  year={2018},
  organization={IEEE}, 
  doi={10.1109/CCWC.2018.8301729}
}

@inproceedings{vgg19,
  title={A comparison between VGG16, VGG19 and ResNet50 architecture frameworks for Image Classification},
  author={Mascarenhas, Sheldon and Agarwal, Mukul},
  booktitle={2021 International conference on disruptive technologies for multi-disciplinary research and applications (CENTCON)},
  volume={1},
  pages={96--99},
  year={2021},
  organization={IEEE},
  doi={10.1109/CENTCON52345.2021.9687944}}

@inproceedings{cnn,
  title={Simple convolutional neural network on image classification},
  author={Guo, Tianmei and Dong, Jiwen and Li, Henjian and Gao, Yunxing},
  booktitle={2017 IEEE 2nd International conference on big data analysis (ICBDA)},
  pages={721--724},
  year={2017},
  organization={IEEE.},
doi={10.1109/ICBDA.2017.8078730}}

@misc{groom2024google,
  author       = {Groom, N.},
  year         = {2024},
  month        = {June},
  day          = {13},
  title        = {Google partners with Nevada utility for geothermal to power data centers},
  howpublished = {Reuters},
  url          = {https://www.reuters.com/business/energy/google-partners-with-nevada-utility-geothermal-power-data-centers-2024-06-13},
}

@misc{dcd2024microsoft,
  author       = {{DCD News}},
  year         = {2024},
  month        = {July},
  day          = {18},
  title        = {Microsoft signs solar PPAs totaling 475MW with AES},
  howpublished = {DataCenterDynamics},
  url          = {https://www.datacenterdynamics.com/en/news/microsoft-signs-solar-ppas-totaling-475mw-with-aes},
}

@misc{google2024sustainability,
  author       = {{Google}},
  year         = {2024},
  title        = {Operating sustainably},
  howpublished = {\url{https://datacenters.google/operating-sustainably}},
  note         = {Google Data Centers},
}

@article{acosta2024generative,
  title={Generative artificial intelligence and web accessibility: Towards an inclusive and sustainable future},
  author={Acosta-Vargas, Patricia and Salvador-Acosta, Bel{\'e}n and Novillo-Villegas, Sylvia and Sarantis, Demetrios and Salvador-Ullauri, Luis},
  journal={Emerging Science Journal},
  volume={8},
  number={4},
  pages={1602--1621},
  year={2024},
  publisher={Emerging Science Journal}, 
doi={https://doi.org/10.28991/ESJ-2024-08-04-021}
}

@article{sookhom2023tourism,
  title={The new way of tourism in green economy style for sustainable community development and empowerment},
  author={Sookhom, A and Krivart, K and Jansiri, N and Sookhom, A and Chanasit, J and Pathak, S and Mallongi, A},
  journal={HighTech and Innovation Journal},
  volume={5},
  number={4},
  pages={1085--1100},
  year={2024}, 
doi={ https://doi.org/10.28991/HIJ-2024-05-04-015}
}

@article{shaari2024carbon,
  title={The carbon conundrum: exploring CO2 emissions, public debt, and environmental policy},
  author={Shaari, Mohd Shahidan and Sulong, Amri and Ridzuan, Abdul Rahim and Esquivias, Miguel Angel and Lau, Evan},
  journal={Emerging Science Journal},
  volume={8},
  number={3},
  pages={933--947},
  year={2024},
  publisher={Ital Publication}, 
doi={ https://doi.org/10.28991/ESJ-2024-08-03-08}
}

@article{Prisma2021,
  title={The PRISMA 2020 statement: an updated guideline for reporting systematic reviews},
  author={Page, Matthew J and McKenzie, Joanne E and Bossuyt, Patrick M and Boutron, Isabelle and Hoffmann, Tammy C and Mulrow, Cynthia D and Shamseer, Larissa and Tetzlaff, Jennifer M and Akl, Elie A and Brennan, Sue E and others},
  journal={bmj},
  volume={372},
  year={2021},
  publisher={British Medical Journal Publishing Group}, 
doi={https://doi.org/10.1136/bmj.n71 }
}

@article{10136725,
  title={Evaluating the potential of wavelet pooling on improving the data efficiency of light-weight cnns},
  author={El-Bana, Shimaa and Al-Kabbany, Ahmad and Elragal, Hassan M and El-Khamy, Said},
  journal={IEEE Access},
  volume={11},
  pages={51199--51213},
  year={2023},
  publisher={IEEE}, 
  doi={10.1109/ACCESS.2023.3280191}}

@article{LecunDL,
  title={Deep learning},
  author={LeCun, Yann and Bengio, Yoshua and Hinton, Geoffrey},
  journal={nature},
  volume={521},
  number={7553},
  pages={436--444},
  year={2015},
  publisher={Nature Publishing Group UK London}, 
  doi={https://doi.org/10.1038/nature14539}
}

@article{Saddami2024AdvancingGA,
  title={Advancing Green AI: Efficient and Accurate Lightweight CNNs for Rice Leaf Disease Identification},
  author={Saddami, Khairun and Nurdin, Yudha and Zahramita, Mutia and Safiruz, Muhammad Shahreeza},
  journal={arXiv preprint},
  year={2024}, 
doi={https://doi.org/10.48550/arXiv.2408.01752}
}

@article{hsu2022language,
  title={Language model compression with weighted low-rank factorization},
  author={Hsu, Yen-Chang and Hua, Ting and Chang, Sungen and Lou, Qian and Shen, Yilin and Jin, Hongxia},
  journal={arXiv preprint},
  year={2022}, 
doi={https://doi.org/10.48550/arXiv.2207.00112}
}

@article{mwangi2014review,
  title={A review of feature reduction techniques in neuroimaging},
  author={Mwangi, Benson and Tian, Tian Siva and Soares, Jair C},
  journal={Neuroinformatics},
  volume={12},
  number={2},
  pages={229--244},
  year={2014},
  publisher={Springer}, 
  doi={10.1007/s12021-013-9204-3}
}

@inproceedings{hernandez2018further,
  title={Further advantages of data augmentation on convolutional neural networks},
  author={Hern{\'a}ndez-Garc{\'\i}a, Alex and K{\"o}nig, Peter},
  booktitle={International Conference on Artificial Neural Networks},
  pages={95--103},
  year={2018},
  organization={Springer.},
  doi={https://doi.org/10.48550/arXiv.1906.11052}
}

@article{wang2025comprehensive,
  title={A comprehensive survey on data augmentation},
  author={Wang, Zaitian and Wang, Pengfei and Liu, Kunpeng and Wang, Pengyang and Fu, Yanjie and Lu, Chang-Tien and Aggarwal, Charu C and Pei, Jian and Zhou, Yuanchun},
  journal={IEEE Transactions on Knowledge and Data Engineering},
  year={2025},
  publisher={IEEE.},
  doi={10.1109/TKDE.2025.3622600}}

@article{ali2014data,
  title={Data normalization and standardization: a technical report},
  author={Ali, Peshawa Jamal Muhammad and Faraj, Rezhna Hassan and Koya, Erbil and Ali, Peshawa J Muhammad and Faraj, Rezhna H},
  journal={Mach Learn Tech Rep},
  volume={1},
  number={1},
  pages={1--6},
  year={2014}, 
  doi={10.13140/RG.2.2.28948.04489}
}

@inproceedings{cabello2023impact,
  title={The impact of data normalization on the accuracy of machine learning algorithms: A comparative analysis},
  author={Cabello-Solorzano, Kelsy and Ortigosa de Araujo, Isabela and Pe{\~n}a, Marco and Correia, Lu{\'\i}s and J. Tall{\'o}n-Ballesteros, Antonio},
  booktitle={International conference on soft computing models in industrial and environmental applications},
  pages={344--353},
  year={2023},
  organization={Springer}, 
doi={https://doi.org/10.1007/978-3-031-42536-3_33}
}

@article{van2009dimensionality,
  title={Dimensionality reduction: a comparative review.},
  author={Van Der Maaten, Laurens and Postma, Eric and Van den Herik, Jaap},
  journal={Journal of Machine Learning Research},
  volume={10},
  number={66-71},
  year={2009}
}

@incollection{trzcinski2024zero,
  title={Zero-waste machine learning},
  author={Trzcinski, Tomasz and Twardowski, Bart{\l}omiej and Zieli{\'n}ski, Bartosz and Adamczewski, Kamil and W{\'o}jcik, Bartosz},
  booktitle={ECAI 2024},
  pages={43-49},
  publisher={IOS Press},
  address={Amsterdam},
  year={2024}, 
doi={10.3233/FAIA240466}
}

@article{patterson2021carbon,
  title={Carbon emissions and large neural network training},
  author={Patterson, David and Gonzalez, Joseph and Le, Quoc and Liang, Chen and Munguia, Lluis-Miquel and Rothchild, Daniel and So, David and Texier, Maud and Dean, Jeff},
  journal={arXiv preprint arXiv:2104.10350},
  year={2021.}, 
  doi={https://doi.org/10.48550/arXiv.2104.10350}
}
\end{document}